\documentclass{article}

\usepackage{arxiv}

\usepackage[utf8]{inputenc} 
\usepackage[T1]{fontenc}    
\usepackage{hyperref}       
\usepackage{url}            
\usepackage{booktabs}       
\usepackage{amsfonts}       
\usepackage{nicefrac}       
\usepackage{microtype}      
\usepackage{comment}
\usepackage{rotating}
\usepackage{xcolor}
\usepackage{graphicx}
\usepackage{natbib}
\usepackage{doi}

\usepackage{amssymb}
\usepackage{pifont}
\newcommand{\cmark}{\ding{51}}%
\newcommand{\xmark}{\ding{55}}%

\title{A Knowledge Graph Embeddings based Approach for Author Name Disambiguation using Literals}

\date{}	

\author{{Cristian Santini}\\
	FIZ Karlsruhe - Leibniz Institute of Technology\\
	Karlsruhe Institute of Technology\\
	Karlsruhe, Germany\\
	\texttt{cristian.santini@fiz-karlsruhe.de} \\
	\And
	Genet Asefa Gesese\\
	FIZ Karlsruhe - Leibniz Institute of Technology\\
	Karlsruhe Institute of Technology\\
	Karlsruhe, Germany\\
	\texttt{genet-asefa.gesese@fiz-karlsruhe.de} \\
	\AND
	Silvio Peroni \\
	Department of Classical Philology and Italian Studies\\
	University of Bologna \\
	Bologna, Italy\\
	\texttt{silvio.peroni@unibo.it} \\
	\And
	Aldo Gangemi \\
	Department of Classical Philology and Italian Studies\\
	University of Bologna \\
	Bologna, Italy\\
	\texttt{aldo.gangemi@unibo.it} \\
	\And
	Harald Sack \\
	FIZ Karlsruhe - Leibniz Institute of Technology\\
	Karlsruhe Institute of Technology\\
	Karlsruhe, Germany\\
	\texttt{harald.sack@kit.edu} \\
	\And
	Mehwish Alam\\
	FIZ Karlsruhe - Leibniz Institute of Technology\\
	Karlsruhe Institute of Technology\\
	Karlsruhe, Germany\\
	\texttt{mehwish.alam@kit.edu} \\
}

\renewcommand{\headeright}{}
\renewcommand{\undertitle}{}
\renewcommand{\shorttitle}{LAND - Literally Author Name Disambiguation}

\hypersetup{
pdftitle={A template for the arxiv style},
pdfsubject={q-bio.NC, q-bio.QM},
pdfauthor={David S.~Hippocampus, Elias D.~Striatum},
pdfkeywords={First keyword, Second keyword, More},
}

\begin{document}
\maketitle

\begin{abstract}
Scholarly data is growing continuously containing information about the articles from a plethora of venues including conferences, journals, etc. Many initiatives have been taken to make scholarly data available in the form of Knowledge Graphs (KGs). These efforts to standardize these data and make them accessible have also led to many challenges such as exploration of scholarly articles, ambiguous authors, etc. This study more specifically targets the problem of Author Name Disambiguation (AND) on Scholarly KGs and presents a novel framework, Literally Author Name Disambiguation (LAND), which utilizes Knowledge Graph Embeddings (KGEs) using multimodal literal information generated from these KGs.  
This framework is based on three components: 1) Multimodal KGEs, 2) A blocking procedure, and finally, 3) Hierarchical Agglomerative Clustering. Extensive experiments have been conducted on two newly created KGs: (i) KG containing information from Scientometrics Journal from 1978 onwards (OC-782K), and (ii) a KG extracted from a well-known benchmark for AND provided by AMiner (AMiner-534K). The results show that our proposed architecture outperforms our baselines of 8-14\% in terms of F$_1$ score and shows competitive performances on a challenging benchmark such as AMiner. The code and the datasets are publicly available through Github (\url{https://github.com/sntcristian/and-kge}) and Zenodo (\url{https://doi.org/10.5281/zenodo.6309855}) respectively.
\end{abstract}


\section{Introduction}
\label{sec:introduction}

Data available in Scholarly Knowledge Graphs (SKGs) -- i.e., ``a graph of data intended to accumulate and convey knowledge of the real world, whose nodes represent entities of interest and whose edges represent potentially different relations between these entities''~\cite{HoganBCdMGKGNNN21} -- is growing continuously every day, leading to a plethora of challenges concerning, for instance, article exploration and visualization in \cite{liu_survey_2018}, article recommendation in \cite{beel_research-paper_2016}, citation recommendation in \cite{farber_citation_2020}, and Author Name Disambiguation (AND) (see \cite{sanyal_review_2021} for a survey), which is relevant for the purposes of the present article. In particular, AND refers to a specific task of entity resolution which aims at resolving author mentions in bibliographic references to real-world people. 

Author persistent identifiers, such as ORCIDs and VIAFs, simplify the AND activity since such identifiers can be used for reconciling entities defined as different objects and representing the same real-world person. However, the availability of such persistent identifiers in SKGs -- such as OpenCitations (OC)~\cite{peroni_opencitations_2020}, AMiner \cite{wan_aminer_2019} and Microsoft Academic Knowledge Graph (MAKG)~\cite{farber_microsoft_2019} -- is characterized by very low coverage and, as such, additional and computationally-oriented techniques must be adopted to identify different authors as the same person.

In the past, many approaches have been developed to automatically address AND by using publications metadata (e.g., title, abstract, keywords, venue, affiliation, etc.) to extract features that can be used in the disambiguation task. These methods vary widely from supervised learning methods to unsupervised learning including recently developed deep neural network-based architectures. However, the existing SKGs do not provide all the relevant contextual information necessary to effectively and efficiently reuse those approaches, which often rely on pure textual data. 

This issue is represented by the fact that only a few bibliographic databases provide extensive access to full bibliographic metadata, including abstracts, keywords and uniquely identified affiliations (e.g. with ROR IDs\footnote{\url{https://ror.org/}}), upon which many current AND approaches heavily rely. As witnessed by the rise of Open Access initiatives such as the Initiative 4 Open Abstracts (I4OA\footnote{\url{https://i4oa.org/}}), the availability of textual metadata such as abstracts, which are used by many machine learning approaches for bibliometric studies (including AND), is limited. As an example, in June 2020, on Crossref, a non-profit DOI registration agency that openly provides high-quality metadata from most international publishers, only 8\% of journal articles had an abstract. This is of course a challenge posed by the infrastructure involved in the provision of SKGs, such as OpenCitations, which usually collect these metadata from heterogeneous and openly available bibliographic databases.

In contrast to many current approaches, this study focuses on performing AND for scholarly data represented as linked data or included in SKGs, by considering the multi-modal information available in such collections, i.e., the structural information consisting of entities (e.g., authors, publications, venues and affiliations) and relations between them as well as text or numeric values associated with the authors and publications defined in the form of literals (family name, given name, publication title, venue title, year of publication). In order to take into account discriminative features, such as abstracts and keywords, we present a novel approach which models structural information encoded in the SKG and exploits the semantics conveyed by potential literals to refine the structural representations.

The proposed framework to address this task is named \emph{Literally Author Name Disambiguation} (LAND), which focuses on tackling the following research questions:

\begin{itemize}
    \item Can Knowledge Graph Embeddings (KGEs) -- i.e. a technique that enables the creation of a ``dense representation of the graph in a continuous, low-dimensional vector space that can then be used for machine learning tasks'' \cite{gesese_survey_2021} -- be used effectively for the downstream task of clustering, more specifically for author name disambiguation?
    \item Does the information present in attributive triples (i.e. titles, publication dates, etc.) in existing SKGs enhance the aforementioned representations for AND?
\end{itemize}

The goal of this article is to provide a representation learning method for extracting entity features from SKGs which do not require any labeled training data. To this end, LAND uses semantic matching models which incorporate literal information, namely LiteralE \cite{kristiadi_incorporating_2019}, to extract author-related features which can adapt to the sparsity of metadata in SKGs. LAND further makes use of KGEs along with Hierarchical Agglomerative Clustering (HAC) and Blocking (for a survey see \cite{backes_impact_2018}) in order to obtain final clusters of authors from the dense representations learned directly from the SKGs.

The rest of the article is organized as follows. Section~\ref{sec:related_work} discusses the related studies in the field. Section~\ref{sec:dataset} introduces the SKGs created for conducting our experiments. Section~\ref{sec:framework} details the proposed framework, while Section~\ref{sec:experiments} documents the conducted experiments and the achieved results. Finally, Section~\ref{sec:conclusion} provides a summary of the work and gives some future perspectives.
\section{Related Work}
\label{sec:related_work}

This section describes the studies related to author name disambiguation which are further divided into supervised learning approaches, unsupervised learning, graph-based approaches. It also details the studies using KGEs for scholarly data. 

\subsection{Author Name Disambiguation}

Generally, name disambiguation, also called entity resolution, is the task of removing duplicates from large noisy databases. This task is affected by ambiguity itself, due to the multiple names which are used to refer to it in database theory: record linkage, deduplication, data matching, instance matching, and data linkage. A first definition of it was provided by \cite{dunn_record_1946}, in which record linkage is defined as the process of collecting pieces of information which refer to the same individual into one single unit. Later, \cite{fellegi_theory_1969} provided a mathematical modelling of the problem.

The relevance of name disambiguation is related to the fact that it is required in many scientific domains: not only statistics but also political sciences and social sciences, medicine and epidemiology, demographic and human rights statistics, bibliometric and scientometric analysis. For a recent interdesciplinary survey please refer to \cite{binette_almost_2022}. In general, the problem of entity resolution can be tackled via two distinct approaches: either clustering records according to the real-world entity to which they belong, or identifying coreferent record pairs.

Author Name Disambiguation (AND) is a subcategory of the name disambiguation task, concerned with the resolution of records inside scholarly databases. In~\cite{ferreira_brief_2012}, the authors classify the existing AND approach into two categories, i.e., \emph{author assignment} and \emph{author grouping}. The author assignment approach directly assigns a label to every item corresponding to a real-world author. This approach is often difficult to implement since it requires the actual list of authors to be mapped to specific classes \emph{a priori}. The second method, author grouping, consists in clustering the entries corresponding to authors via a similarity function which should produce output groups associated with real-world entities. Author grouping may not require the number of authors to be known \emph{a priori} and is consequently easier to implement in most cases. 

Moreover, in the aforementioned survey, the authors classify the evidence used according to three categories: 1) \emph{web information} (e.g., information extracted from web pages), 2) \emph{citation information} (i.e. metadata associated with publications), and 3) \emph{implicit evidence}, such as topic modeling or graph embeddings. Additionally, a common strategy in AND is to initially group author entries into subsets by looking at name compatibility, e.g., authors carrying the same last name are grouped and disambiguation is performed within each group. This activity is carried out to reduce the number of pairwise comparisons required by the task and is termed as Blocking (for details see \cite{backes_impact_2018}). One of the simplest and most common approaches is to group authors by looking at the full last name and the first initial (hereafter, \emph{LN-FI}) of the given name, therefore called LN-FI blocking.

\subsubsection{Supervised Learning Approaches}

These approaches take into consideration several features extracted from scholarly metadata, such as publications' title words, keywords, coauthors, venues, etc., and a classifier is trained with labeled data pairs to estimate if a given pair of publications belong to the same author or not.  One of the seminal works in supervised AND is Author-ity, presented in \cite{torvik_author_2009}. Author-ity makes use of LN-FI blocking to preliminarily split publications related to ambiguous author names into blocks; then, given a pair of publications $p_1, p_2$ corresponding to two author name instances $s_1, s_2$ respectively in a block, it constructs a multidimensional similarity profile $x(p_1, p_2)$, based on title, journal name, co-author names, MeSH, language, affiliation, email, and other name attributes. The similarity profile is the input feature of a classifier trained with Bayesian learning to estimate the probability of $x(p_1, p_2)$, given that $p_1, p_2$ are written by the same author or not. In the end, a maximum-likelihood-based agglomerative clustering is used in order to group publications. 

Another approach that makes use of supervised learning is BEARD, presented in \cite{ngonga_ngomo_ethnicity_2016}. This method adopts a phonetic-based blocking strategy to preliminarily group authors into blocks by looking at the phonetic representation of the normalized surname (e.g., ``van der Waals, J. D." to “Waals, J. D.”). Moreover, it associates a set of features to each pair of author instances that are designed to be sensitive to the ethnic group of the authors. Then, a classifier is trained on annotated data to learn a pairwise distance function using tree-based methods (i.e. Random Forest (RF) and Gradient Boosted Trees (GBT)). Finally, author references are grouped using hierarchical agglomerative clustering. The novelty of this method is to introduce for the first time ethnicity-sensitive features to make author name disambiguation sensitive to the actual origin of the authors. However, the authors state that the impact of the phonetic-based blocking strategy is not adequately addressed.

In \cite{tran_author_2014}, a Deep Neural Network (DNN) is used to learn distinctive features for author disambiguation. By using labeled pairs of publications related to an ambiguous author name, the neural network is trained to calculate the posterior probability of the internal features, given basic features as input, such as the Jaccard distance between author name, co-authors list, affiliation, keywords, and author interest keyword.

In \cite{kim_hybrid_2019}, the authors exploit structure-based features, consisting of word embeddings modeling publications' metadata using Term Frequency-Inverse Document Frequency (TF-IDF), to train four distinct architectures, a Support Vector Machine (SVM), an RF, a GBT, and a DNN. These architectures are trained to determine whether a pair of publications is related to the same author or not.

\subsubsection{Unsupervised Learning Approaches}
Due to the fact that human annotation of training data is a time-consuming task for digital libraries, many unsupervised approaches have been devised for AND. One of the seminal works in unsupervised methods was presented in \cite{cota_unsupervised_2010}, where the authors propose an unsupervised heuristic-based AND approach. In their approach, publications related to an ambiguous name are clustered together based on shared coauthors. Subsequently, the fragmented clusters are linked together based on the cosine similarity between TF-IDF vectors of potentially weak features such as title and publication venue. However, the drawbacks of this approach are mainly due to the fact that TF-IDF vectors do not capture efficiently similarity between titles. Due to this reason, this method heavily relies on coauthorship information. This may lead to false positives and data segmentation, especially when an author does not have the same set of coauthors across multiple works.

Similar to \cite{cota_unsupervised_2010}, the methodology presented \cite{liu_fast_2015} consists of an incremental clustering approach where the authors rely on coauthorship information to create fragmented clusters, which are later merged by using similarity in title and publication venue. More specifically, in the first step, coauthorship-based clusters are created by using  the Floyd-Warshall algorithm; then, in the second step, clusters are merged by using cosine similarity between publication titles; in the last step relations between authors and venues are captured by using Latent Semantic Analysis (LSA) to further merge fragmented clusters.

In \cite{caron_large_2014}, the authors adopt a predefined set of rules for scoring author similarity by taking into consideration several metadata attributes of a pair of publications: for example, if one paper cites the other it gets a score of 10. If the overall score of a pair of publications is above a certain threshold, the two publications are considered as belonging to the same author. The results obtained after pairwise comparison are then aggregated using HAC.  For its simplicity and interpretability, this method was recently adopted effectively in \cite{farber_microsoft_2022} to perform large-scale author name disambiguation on MAKG \cite{farber_microsoft_2019}. However, despite its scalability, this method does not take into account higher-order relations between entities in a dataset, such as a coauthorship and citation patterns, since it only takes into account the amount of shared information between the metadata of the publications. Moreover, the use of predefined rules might not be effective for author name disambiguation on datasets affected by data sparsity and therefore might not be flexible to different domains and data sources.

Recently in \cite{waqas_multilayer_2021}, a heuristic-based clustering framework was proposed to combine the information coming from structure-based features (e.g. metadata such as title, coauthor names, venues, etc.) with global features, which are distributional representations obtained using word embeddings, to perform AND in a multi-step algorithm. Similar to other studies, the authors create initial clusters by using more discriminative features, such as co-authors, author affiliation, and author email. In order to compare the information present in the structure-based features, the authors apply several heuristic rules which vary with respect to the information considered. In the subsequent step, the algorithm merges the obtained clusters by using similarity between word embeddings extracted from the stemmed words in titles, abstracts, and keywords. In the end, the authors use venues and publication dates to merge together unwanted split clusters. As for other heuristic-based methods, this approach does not capture higher-order relations between publications, such as citation networks; moreover, the use of a multi-layer approach does not allow to improve precision as further information is considered, such that coming from titles and venues, which is only used to merge wrongly split clusters.

\subsubsection{Graph-based Approaches}
In \cite{fan_graph-based_2011}, a graph-based method called GHOST is exploited to cluster publications using graph components. In this method, a co-authorship graph is constructed for each instance \emph{s} related to a queried author name by collapsing all the co-authors with same name into one single node. The resulting graph contains all authors which are co-authors with \emph{s} and all authors which have co-authored a paper with the co-authors of \emph{s}. 
Then, the similarity between two instances of \emph{s} is computed based on the number of valid paths and affinity propagation is used to group nodes into clusters. However, this method does not work for single-author papers, and information contained in other metadata (e.g. publication titles, abstracts, or keywords) is not considered.

In \cite{km_graph_2020}, the authors present a graph-based approach where two graphs are combined together, a person-person graph obtained by connecting papers with shared coauthors, and a document-document graph which models similarity between publications' content. The document-document graph is obtained by first modeling abstract keywords with TF-IDF vectors and by drawing an edge between two nodes of the graphs when their similarity is higher than a selected threshold; subsequently, this graph is pruned by removing connections between papers whose shared referenced works are below a certain threshold. Then the person-person and the document-document graphs are merged together and the connected components of the combined graph represent the final authors. This graph-based model takes into account citation information from the two papers, however, the construction of the person-person graph might produce false split clusters in datasets from the humanities domain, where the same coauthors might not appear in many papers.

\paragraph{Graph Embedding based Approaches}
\label{sec:graph_embeddings}

Recently, graph embeddings on heterogeneous and non-heterogeneous graphs have been implemented to solve the AND problem. In \cite{zhang_name_2017}, the authors propose an AND approach that works on anonymized graphs by using relational information learned via network embeddings. This method constructs three local graphs for a candidate set of documents: a person-person graph representing a collaboration between authors, a person-document graph representing the association between authors and bibliographic records, and a document-document similarity graph based on co-authorship relations. A representation learning model is proposed to embed the nodes of these graphs into a shared low-dimensional space by optimizing a joint objective function based on the pairwise ranking of similarity. The final results are generated by HAC. This method proposes a new representation learning framework that is particularly suited for downstream clustering tasks. However, since this approach is designed for anonymized graphs, it does not take into consideration many attributes of nodes instead it considers co-author sharing for computing document similarity.

A network learning algorithm that models different types of relationships among publications, called Diting is proposed in \cite{peng_diting_2019}. This approach first builds different weighted networks for each paper belonging to an ambiguous name, where each weighted network represents the similarity of the publications in a candidate set with respect to a selected feature, such as title, coauthor, organization, venue, etc. Then, network embeddings are learned by maximizing the gap between similar nodes and dissimilar ones, and graph coarsening is used to reduce the number of nodes. Final authors are obtained by using HDBSCAN and Affinity Propagation (AP) to cluster the node embeddings. In the same paper, the authors propose a semi-supervised version of their method, called Diting++, which makes use of information available on the web to refine the final author's clusters.

In \cite{zhang_name_2018}, the authors propose an AND method based on three components: a global learning module which creates embedding representations for each document by leveraging structure-based features, a local-linkage learning framework which exploits shared information to refine the embeddings related to an ambiguous name \emph{a}, and a Recurrent Neural Network (RNN) which estimates the number of clusters for each ambiguous name \emph{a}. In the global learning step, the authors refine structure-based features obtained from the metadata about the publication by using labeled triplet samples. In the second step, these global features are refined after modeling the candidate set of publications related to an ambiguous author name in a weighted network and by using a graph autoencoder to learn node embeddings. The final partition of the candidate set is obtained by using HAC. Despite the high performance of this model, its reproducibility is hindered by the fact that it requires labelled samples and complex feature engineering, which can make the approach not flexible to data sparsity.

In \cite{zhang_author_2019}, a simple graph embedding method based on Random Walk and Skip-Gram \cite{mikolov_efficient_2013} is proposed to perform AND on co-authorship networks. \cite{qiao_unsupervised_2019} presents an AND methodology for heterogeneous weighted graphs. In this method, publications are represented in vector space by using Doc2Vec~\cite{le_distributed_2014}; then, publication embeddings are refined by using information from a heterogeneous network that links publications based on different weighted relations (e.g., CoTitle, CoVenue, CoAuthor). The information from this network is used to refine the Doc2Vec embeddings by using a Heterogeneous Graph Convolutional Neural Network (HGCNN).

In \cite{wang_author_2020}, an author disambiguation algorithm for heterogeneous information networks is presented. In heterogeneous information networks, relations between scholarly works, authors, and other entities are modelled with nodes and relations of different types. In their method, the authors model the content of the article coming from literary attributes such as titles and abstracts with Doc2Vec, in order to create a document embedding for each paper; moreover, they encode the heterogeneous information network of scholarly publications into node embeddings by using Node2Vec \cite{grover_node2vec_2016}. The novelty of their method consists in using Generative Adversarial Networks (GANs) to mutually exploit the information coming from document embeddings and node embeddings. By evaluating their method on the AMiner benchmark \cite{zhang_name_2018}, they outperformed the best model by +5.13\% in Macro-F$_1$ score. In their paper, the authors also test the approach on a SKG called Ace-KG, however, the lack of a source code makes it difficult to reproduce this methodology on other datasets.

\cite{pooja_exploiting_2021} proposes an unsupervised graph embedding method to solve the AND task which exploits similarities between publications on two different dimensions: on the content level (e.g. title, abstract, keywords, references) and on the coauthor level. In order to do so, they use two different variational graph autoencoders and neural fusion to combine multiple representations.

In \cite{chen_name_2021}, the authors propose an author disambiguation algorithm based on Graph Convolutional Networks (GCNs) which takes into account attribute features as well as relation information. In their method, they first build three graphs: a paper-paper, a person-person and a paper-person graphs, which are then fed to a specialized GCN that outputs hybrid representation. The authors show that their method performs in a competitive way on multiple datasets, including AMiner \cite{zhang_name_2018}.

A detailed comparison of different graph embedding models, along with the information used and the respective performances on different benchmark datasets is presented in \emph{Table \ref{tab:table_graph_and}}. As shown in the table, the majority of these methods do not apply representation learning directly on graphs but they preliminary require feature engineering to measure the relatedness of a pair of publications. Moreover, only \cite{chen_name_2021}, \cite{wang_author_2020}, and \cite{zhang_name_2017} are tested on a scenario where few publication attributes are used (only titles, coauthors, venue, and affiliation) and other textual information, like abstracts, is not considered.

\begin{sidewaystable}[]
    \centering
       \caption{Overview of graph embedding AND methods}
    \label{tab:table_graph_and}
    \resizebox{\textwidth}{!}{
    \begin{tabular}{|p{4cm}|p{5cm}|p{5cm}|p{4cm}|p{4cm}|p{4cm}|p{2cm}|}
    \hline 
         \textbf{References} & \textbf{Technique} & \textbf{Information used} & \textbf{Dataset} & \textbf{Results} & \textbf{Feature engineering} & \textbf{Supervised}\\ \hline
         \cite{zhang_name_2017} & Network embedding on anonymized graphs + HAC & author network & AMineR-2018 & Macro-F1: 62.81 & \cmark & \xmark \\ \hline
         \cite{peng_diting_2019} & Network embedding with margin loss + HDBSCAN & title, coauthor, venue, organization, summary & AMiner-2012, DBLP, CiteSeerX & Macro-F1: 74.5 (AMiner-2012), Macro-F1: 83.2 (DBLP), Macro-F1: 63.5 (CiteSeerX) & \cmark & \xmark \\ \hline
         \cite{zhang_author_2019}& Network embedding on anonymized graphs + HAC & author network & 10 ambiguous names from CiteSeerX & Macro-F1: 62.13 & \xmark & \xmark\\ \hline
         \cite{qiao_unsupervised_2019}& Doc2Vec + Heterogeneous Graph CNN + HAC & title, abstract, keywords, coauthors, venue & Aminer-2012 & Macro-F1: 78.6 & \cmark & \xmark\\ \hline
         \cite{zhang_name_2018} & Supervised global learning + unsupervised local graph encoder + cluster size estimation with RNN + HAC & title, abstract, coauthors, venue, affliations, keywords &  AMiner-2018 & Macro-F1: 67.79 & \cmark & \cmark \\ \hline
         \cite{wang_author_2020} & Document embeddings with word2vec model + Adversarial Representation Learning + HAC & titles and abstracts, coauthors, organization, venue, keywords & AMiner-2018 & Macro-F1: 72.92 & \xmark & \xmark \\ \hline
         \cite{pooja_exploiting_2021}& Variational Graph Autoencoder + Neural fusion + HAC& title, abstract, keywords, venue, affiliation, coauthors & AMiner-2012, Aminer-2018 & Macro-F1:78.4 (AMiner-2012), Macro-F1: 66.9 (AMiner-2018)& \cmark & \xmark\\ \hline
         \cite{chen_name_2021} & Document embeddings with word2vec model + Graph Convolutional Neural Network + HAC & title, keywords, venue, coauthor, organization & AMiner-2018 & Macro-F1: 65.71 & \cmark & \xmark \\ \hline
    \end{tabular}
    }
 
\end{sidewaystable}

\subsection{Knowledge Graph Embeddings and Scholarly Data}
Few studies have been made recently on the use of KGEs with an  application on scholarly linked data. In \cite{mai_combining_2018}, an entity retrieval system for the scholarly domain was proposed, combining information coming from textual embeddings and structural embeddings trained from the KG IOS Press LD Connect\footnote{\url{http://ld.iospress.nl/}}. In this paper, the authors evaluate the quality of low-dimensional representations of papers and entities (i.e. authors, organizations, etc.) by extracting two benchmark datasets: 1) a benchmark dataset collected from Semantic Scholar in order to evaluate the semantic similarity of papers, and 2) a second benchmark dataset extracted from DBLP used in order to evaluate co-authorship recommendations based on KGEs. The authors extract paragraph vectors for representing papers’ content by using Doc2Vec and train TransE \cite{bordes_translating_2013} for extracting embeddings of entities in the SKG of IOS LD Connect. In order to build the entity retrieval model, a logistic regression model which takes as input features both paragraph vectors and structural embeddings. It is trained on a dataset of similar papers automatically collected from Semantic Scholar. Reported results show that KGEs do not have a significant impact on paper similarity classification, whether textual embeddings alone achieve robust results. As a second step, a co-author inference evaluation is carried out by using a benchmark dataset extracted from DBLP to demonstrate the ability of TransE for predicting coauthorship links based on the observed triples. 

In \cite{nayyeri_embedding-based_2020}, embeddings have been used to generate coauthorship recommendations on SKGs. One of the aims of this work is to propose a novel approach for training KGE models on SKGs where 1-to-N, N-to-1, and N-to-N relations are frequent (i.e. authorship relations or citation links). In order to address this issue, the authors present a reimplementation of TransE \cite{bordes_translating_2013} and RotatE \cite{sun_rotate_2019} by using a newly proposed loss function optimized for many-to-many relations, i.e. Soft Margin (SM) loss. The results of their study show how the models equipped with SM loss outperform the original models. The novelty of this study is to propose a loss function that mitigates the adverse effects of false-negative sampling and to investigate the use of KGEs for co-authorship suggestions. 
\section{Creation of the Scholarly KGs}
\label{sec:dataset}

This section introduces the benchmark datasets \textbf{OC-782K} and \textbf{AMiner-534K} which are created for evaluating the LAND framework. OC-782K is a subset of the Scientometrics Knowledge Graph available from \cite{massari_bibliographic_2021}, which is built in compliance with the OpenCitations Data Model (OCDM)~\cite{daquino_opencitations_2020}. On the other hand, AMiner-534K is a KG generated from a well-established benchmark dataset\footnote{\url{https://static.aminer.cn/misc/na-data-kdd18.zip}} for AND made available by AMiner in the AND paper \cite{zhang_name_2018}. The two KGs are available on Zenodo\footnote{\url{https://doi.org/10.5281/zenodo.6309855}} at  \cite{santini_dataset_2021} in order to allow the reproducibility of the studies herein described.

\subsection{The OC-782K Knowledge Graph}
In this paper, the Scientometrics KG from \cite{massari_bibliographic_2021} is referred to as \emph{Scientometrics-OC}. This publicly available KG contains bibliographic information about the articles published by the journal Scientometrics\footnote{\url{https://www.springer.com/journal/11192}} from 1978 to the present, along with bibliographic information of all the academic works cited by the articles published by that journal. The dataset named OC-782K is created from Scientometrics-OC by modeling entities related to authors, publications, and venues with different data models suited for the task of AND. 

\begin{figure}
    \centering
    \includegraphics[width=8cm]{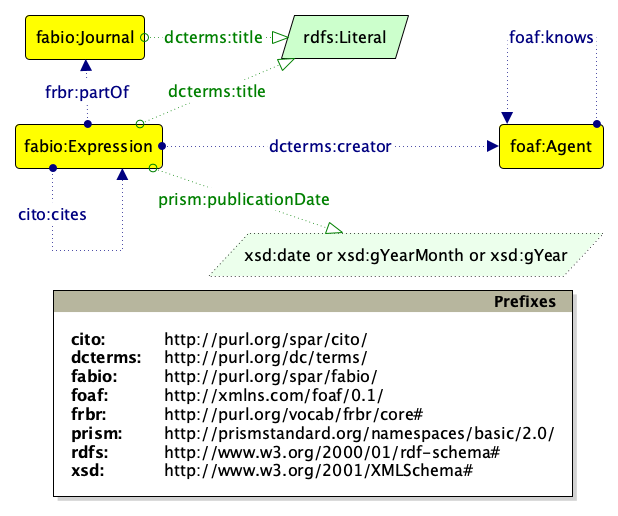}
    \caption{A Graffoo diagram \cite{falco_graffoo_2014} describing the data Model used for OC-782K.}
    \label{fig:datamodel_oc}
\end{figure}

This data model contains three types of entities: {\tt fabio:Expression}, which represents articles, books, conference papers, and other academic works, {\tt fabio:Journal} for representing journal venues (if the related {\tt fabio:Expression} is a journal article), and authors which are described as {\tt foaf:Agent}. The data model is an abstraction of the OCDM \cite{daquino_opencitations_2020} and is created for two reasons: i) for collecting triples only related to the entities of interest (e.g. bibliographic resources, venues, and authors), ii) create an abstract representation of Scientometrics-OC in order to perform representation learning more efficiently. The data model of OC-782K is represented in \emph{Figure ~\ref{fig:datamodel_oc}}. 

OC-782K is extracted from Scientometrics-OC by first collecting information about the bibliographic resources with at least a title and an author. Then, the publication dates and journal venues of these works (if available) were collected. A {\tt foaf:knows} relation is added between two authors who have co-authored the same work, and the relation between two bibliographic resources, a citing expression and a cited one, is represented with the \texttt{cito:cites} relation. Attributive triples were extracted along with triples connecting different entities by following part of the guidelines presented in \cite{DBLP:conf/semweb/GeseseAS21}.

The dataset consists of 781,917 triples, with 620,321 structural triples (i.e. triples with object relations). In the original Scientometrics-OC, while duplicate bibliographic resources and journals were merged by using the DOIs associated with each article, authors are not disambiguated (i.e., there is one author for each {\tt dcterms:creator} relation.) Statistics of the dataset are reported in \emph{Table ~\ref{tab:table_oc_1}} and \emph{Table ~\ref{tab:table_oc_2}}. 

\begin{table}[]
    \centering
       \caption{The number of entities and triples in OC-782K.}
    \label{tab:table_oc_1}
    \begin{tabular}{|c|c|c|c|}
    \hline 
         Object triples & Textual triples & Numeric triples & Entities \\ \hline
     620,321 & 104,621 & 56,975 & 293,186\\ \hline
    \end{tabular}
 
\end{table}

\begin{table}[]
    \centering
       \caption{The number of entities and relations is counted by type in OC-782K.}
    \label{tab:table_oc_2}
    \begin{tabular}{|c|c|c|c|c|c|c|}
    \hline 
        \multicolumn{3}{|c|}{Entities} & \multicolumn{4}{|c|}{Relations} \\ \hline
         Publications & Venues & Authors & dc:creator & foaf:knows & cito:cites & frbr:partOf \\ \hline
      57,266 & 47,355 & 188,565 & 188,565 & 253,942 & 128,738 & 49,076\\ \hline
    \end{tabular}
 
\end{table}

\subsection{The AMiner-534K Knowledge Graph}
\label{sec:aminer-kg}

\begin{figure}
    \centering
    \includegraphics[width=8cm]{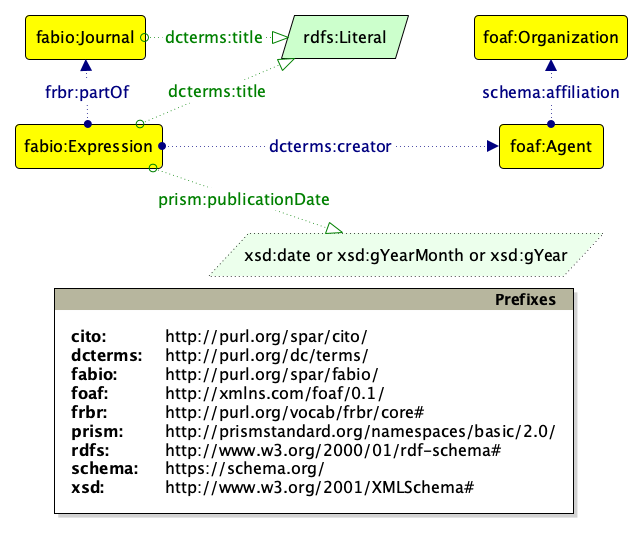}
    \caption{A Graffoo diagram \cite{falco_graffoo_2014} describing the data model used for AMiner-534K.}
    \label{fig:datamodel_aminer}
\end{figure}

In order to evaluate the generalizability of the proposed approach on a different dataset, a second scholarly KG named AMiner-534K is extracted from the AMiner AND benchmark dataset introduced in \cite{zhang_name_2018}. The AMiner benchmark for AND contains a sub-set of publications from AMiner and sampled from 100 ambiguous Asian names. This dataset contains the following information for each scholarly article: title, publication date, venue, keywords, authors, and affiliations. 
The AMiner-534K KG is created by extending the data model of OC-782K with the additional author affiliation information (using the property \texttt{schema:affiliation}). A representation of the data model is available in \emph{Figure \ref{fig:datamodel_aminer}}. However, for AMiner-534K the \texttt{cito:cites} and the \texttt{foaf:knows} properties are absent since these properties are not present in the original benchmark.

Moreover, we did not add abstracts and keywords, even though they are present in the benchmark \cite{zhang_name_2018}. This is due to the fact that this information is often not given in many of the available scholarly datasets and using this information to train a model may lead to results which are not reproducible on datasets not as rich as the benchmark in \cite{zhang_name_2018}. Statistics of the dataset are reported in \emph{Table \ref{tab:aminer_1}} and \emph{Table \ref{tab:aminer_2}}.

\begin{table}[]
    \centering
       \caption{The number of entities and triples in AMiner-534K.}
    \label{tab:aminer_1}
    \begin{tabular}{|c|c|c|c|}
    \hline 
         Object triples & Textual triples & Numeric triples & Entities \\ \hline
     428,473 & 70,046 & 35,021 & 179,377\\ \hline
    \end{tabular}
 
\end{table}

\begin{table}[]
    \centering
       \caption{The number of entities and relations is counted by type in AMiner-534K.}
    \label{tab:aminer_2}
    \begin{tabular}{|c|c|c|c|}
    \hline 
        \multicolumn{4}{|c|}{Entities} \\ \hline
    Publications & Venues & Authors & Organizations \\ \hline
     35,023 & 5,889 & 110,837 & 27,628 \\ \hline
    \multicolumn{4}{|c|}{Relations} \\ \hline
    dc:creator & \multicolumn{2}{|c|}{schema:affiliation} & frbr:partOf  \\ \hline
    197,249 & \multicolumn{2}{|c|}{196,201} & 35,023 \\ \hline
    \end{tabular}
 
\end{table}

\section{Literally Author Name Disambiguation (LAND)}
\label{sec:framework}

In this section, the different components of the proposed framework, Literally Author Name Disambiguation (LAND), are described in detail. \emph{Figure~\ref{sec:framework}} shows the overall architecture of the approach which is based on three main components:

\begin{itemize}
    \item {\bf Multimodal KG Embeddings.} This strategy is aimed at learning representative features of entities and relations in a KG by taking into account the structure of the graph itself along with the semantics contained in the literal about these entities (e.g., titles of academic works or publication dates).
    \item \noindent {\bf Blocking.} This strategy is used to reduce the number of pairwise comparisons required by the AND task by initially grouping authors into blocks characterized by name similarity, so that disambiguation is carried within each block. LAND uses a rather simple but effective blocking strategy called LN-FI blocking.
    \item {\bf Clustering.} Hierarchical Agglomerative Clustering (HAC) is used to group the embeddings associated with each author to be disambiguated into k-clusters by using a vector-based similarity measure (e.g., cosine) along with a distance threshold. 
\end{itemize}

The output of these components is then used for refining the original KG. In the following, each of these components is discussed in detail.


\begin{figure}
    \centering
    \includegraphics[width=\textwidth]{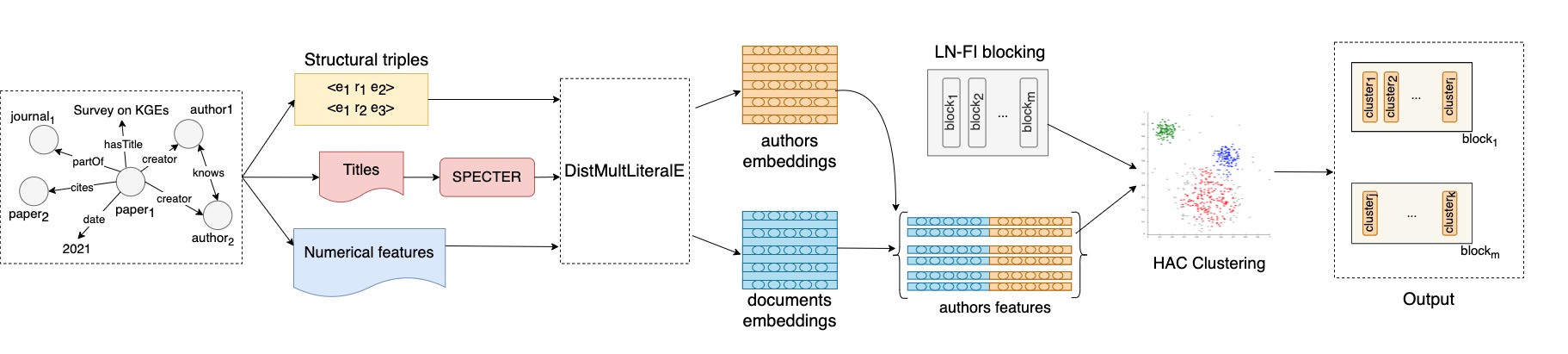}
    \caption{The overview of the LAND architecture}
    \label{fig:pipeline}
\end{figure}


\subsection{Multimodal Knowledge Graph Embeddings}
The first step of the LAND framework is to learn the latent representation of the KGs described in Section~\ref{sec:dataset} including the representations of the authors. To this end, the Multimodal KGEs component of LAND is designed to learn embeddings of entities and relationships in a KG by combining the structural information and literal associated with the entities such as a string or a date value. LAND adopts LiteralE~\cite{kristiadi_incorporating_2019} embedding model in this component to learn the KGEs. It incorporates literal information into entity representations by using a learnable mapping function where the literals can either be numeric or text. 

More specifically, LiteralE is a multimodal extension of \emph{semantic matching models} for learning KGEs, such as DistMult \cite{yang_embedding_2015}. DistMult scores each triple in the KG with a simple bilinear transformation $f(h,r,t)=\textbf{h}^Tdiag(\textbf{r})\textbf{t}$. The representations are then learned during a training phase which is aimed to maximize the score of existing triples in the KG and by minimizing the score of non-existing triples, i.e. negative examples. Meanwhile, LiteralE aims to modify the scoring function $f$ by using for the head $h$ and tail $t$ entities of each triple hybrid representations which combine the information coming from the entities' relations with the information coming from their literal values. At the core of this method is the mapping function $g: R^h \times R^d  \rightarrow R^h$ which takes as input an entity embedding $\textbf{e} \in R^h$  and a literal vector $\textbf{l} \in R^d$ and maps them to a new embedding of the same dimension as the entity embedding.

LAND makes use of SPECTER \cite{cohan_specter_2020}, a pre-trained BERT \cite{devlin_bert_2019} language model for scientific documents in order to encode the textual attributes of the entities (e.g., publication titles) in the vector space $R^d$  before incorporating them into entity vectors with the $g$ function. Each title in our scholarly KG is mapped to a 768-dimensional sentence embedding by utilizing this model. Meanwhile, the numeric datatypes such as \emph{xsd:gYear} are converted to a literal vector as described in LiteralE. 

In this study, the following two varieties of the DistMultLiteralE model are used and compared against their corresponding base (unimodal) model DistMult.

\begin{itemize}
    \item \textbf{LAND-$g_{lin}$}. This architecture incorporates textual embeddings extracted from the titles of the entities (scholarly articles) into their representations by means of a linear transformation defined as follows:
    
    $$g_{lin}(\textbf{e},\textbf{l}) = \textbf{W}[\textbf{e},\textbf{l}] ,$$
    
    where $\textbf{e} \in R^h$ is the vector associated to the $i$th entity in a KG, $ \textbf{l} \in R^d $ is the title embedding, $\textbf{W} \in R^{(h,d+h)}$ is a linear transformation matrix and $[\textbf{e},\textbf{l}] \in R^{(h+d)}$ is the concatenation vector of the entity embedding $\textbf{e}$ and the literal embedding $\textbf{l}$. With this operation, the entity vector is combined with the textual vector encoded by SPECTER into a multi-modal representation of the same dimensionality of the original entity representation $\textbf{e}$.

	\item \textbf{LAND-$g_{gru}$}. The goal here is to leverage both text (titles) and numeric literals (publication dates). This architecture combines the information coming from numeric and textual literals into the entity representations by means of a Gated Recurrent Unit (GRU), defined as follows:
	
	$$g_{gru}(\textbf{e},\textbf{l},\textbf{n})=\texttt{z} \circ \texttt{h} +(1-\texttt{z})\circ \textbf{e}$$
	$$\texttt{z}= \sigma(\textbf{W}_{ze} \textbf{e} + \textbf{W}_{zl} \textbf{l} + \textbf{W}_{zn} \textbf{n} +\textbf{ b})$$
	$$\texttt{h}=h(\textbf{W}_h[\textbf{e},\textbf{l},\textbf{n}]), $$

	where $\circ$ is the element-wise multiplication, $\sigma(\cdot)$ is the sigmoid function, $\textbf{W}_{ze} \in R^{(h,h)},  \textbf{W}_{zl} \in R^{(h,h+d)}, \textbf{W}_{zn} \in R^h$ and  $\textbf{W}_h \in R^{(h,h+d+1)}$ are linear transformation matrices, $\textbf{b}$ is a bias vector, $h(\cdot)$ is a component-wise nonlinearity (e.g. the hyperbolic tangent) and $[\textbf{e},\textbf{l},\textbf{n}]$ is the concatenation of the entity vector, the textual vector and the numeric literal value. With this operation, the entity vector is combined with the textual vector and the numeric literal associated to the publication date into a multi-modal representation of the same dimensionality of the original entity vector $\textbf{e}$.
\end{itemize}

Finally, after having each model trained on a given KG, every author $A$'s  embedding \textbf{E} is modified by concatenating it with the embedding \textbf{D} of the document $D$ (i.e. scholarly article) associated to the author $A$, in order to obtain feature $\textbf{F}$ where $\textbf{F}=\textbf{E}+\textbf{D}$. This is carried out to reflect both the structural information of the two entities (the author and the document) and the literal information present in the document attributes (i.e. title and publication date) in the embedding of the author.

\subsection{Blocking and Clustering}
Blocking is a strategy that is widely used in AND systems. The idea is to split the set of features $F$ related to authors into separate groups, also called \emph{blocks} denoted by $F_{b_1}, F_{b_2},\dots,F_{b_n}$, each one associated with an ambiguous name, so that AND is carried out independently within these blocks. LAND uses the common LN-FI blocking, which divides the set of author features into blocks by looking at the full last name and the first initial of the given name of each author. This blocking technique is chosen since it’s computationally less expensive than other blocking approaches which are based on distance measures or string normalization and it's also compliant with the way publishers often mention author names in the metadata of the publications.

The clustering algorithm in LAND helps in grouping together the author features in each block $F_{b_i}$ into $k$-clusters $\{C_1, ... C_k\}$ where all the features in $C_j$, where $j = {1,...,k}$, ideally belong to the same real-world author. HAC \cite{ward_hierarchical_1963} is used which builds clusters of features in a bottom-up manner. The approach conceives each embedding in a block as a singleton cluster and works by iteratively merging the two most similar clusters until all features have been merged in one final cluster. 

Even though it has been studied that HAC suffers from scalability issues in \cite{on_scalable_2012}, we decided to adopt this clustering method for two main reasons: (\emph{i}) for its simplicity; (\emph{ii}) for the fact that it is the most common clustering method used by graph embedding AND approaches (see \emph{Section \ref{sec:graph_embeddings}}) and allows us to compare LAND with other methods tested on the AMiner benchmark, such as \cite{zhang_name_2017}, \cite{zhang_name_2018}, \cite{wang_author_2020}, \cite{pooja_exploiting_2021} and \cite{chen_name_2021}.

In our implementation of HAC, similarity among clusters is computed with a \emph{single linkage} strategy which, at each step, merges the clusters whose two closest members have the smallest distance, based on cosine similarity. In order to get the final clusters, a threshold on the maximum distance is defined and clusters above this threshold are considered to be corresponding to different authors. The threshold is defined globally over all the blocks by testing different values over the evaluation blocks, more details are provided in \emph{Section \ref{sec:model_selection}}.
\section{Experimental Results}
\label{sec:experiments}

This section discusses the empirical evaluation of the LAND framework. It first shows how the ground truth is generated for the task of AND, then it presents the achieved experimental results of LAND on the newly generated dataset OC-782K and on a KG extracted from a widely used AND benchmark, i.e. AMiner-534K (refer to \emph{Section~\ref{sec:dataset}} for more details). In addition, an error analysis is carried out for the results on OC-782K.

\subsection{Generation of the Ground Truth}
\label{sec:ground_truth}

In order to obtain the ground truth for testing LAND on OC-782K, a list of $(author, ORCID iD)$ pairs is extracted. This is performed for the purpose of having an evaluation dataset of scholarly articles labeled with a unique identifier associated with their real-world authors. 
In order to handle the unbalance in the dataset, only those authors whose last name and first initial are associated with at least two different ORCID iDs are considered. 
The final evaluation dataset contains 630 bibliographic works organized into 184 blocks and 497 different ORCID iDs. Sizes of the largest blocks in this ground-truth are reported in \emph{Table \ref{tab:table_blocks_oc}}.

\begin{table}[]
    \centering
       \caption{10 largest blocks on OC-782K evaluation dataset}
    \label{tab:table_blocks_oc}
    \begin{tabular}{|c|c|}
    \hline 
         \textbf{Name} & \textbf{\# of records} \\ \hline
         Liu W & 28 \\ \hline
         Park H & 27\\ \hline
         Wang X & 27 \\ \hline
         Kim J & 23 \\ \hline
         Zhang L & 15 \\ \hline
         Zhou Y & 14 \\ \hline
         Cabanac G & 13 \\ \hline
         Cai X & 12 \\ \hline
         Kong X & 12 \\ \hline
         Wang J & 12 \\ \hline
    \end{tabular}
 
\end{table}

For measuring the generalizability of the proposed approach, another manually-labeled benchmark dataset is used, i.e., AMiner-534K. This evaluation dataset is larger than the one extracted for OC-782K, with 35,023 scholarly articles and 6,395 unique authors. As for the previous dataset, each ambiguous name is considered a block, and disambiguation is performed within each block.  This evaluation dataset contains 35,129 records divided for 100 ambiguous Asian names. In this dataset, disambiguation is carried independently for each ambiguous name. Sizes of the largest blocks in this ground-truth are reported in \emph{Table \ref{tab:table_blocks_aminer}}.

\begin{table}[]
    \centering
       \caption{10 largest blocks on AMiner-534K evaluation dataset}
    \label{tab:table_blocks_aminer}
    \begin{tabular}{|c|c|}
    \hline 
         \textbf{Name} & \textbf{\# of records} \\ \hline
         Lei Song & 879 \\ \hline
         Yun Zhou & 834\\ \hline
         Jia Xu & 780 \\ \hline
         Lin Huang & 749 \\ \hline
         Song Chen & 744 \\ \hline
         Ping Sun & 742 \\ \hline
         J Yu & 726 \\ \hline
         Yang Shen & 700 \\ \hline
         Xu Xu & 699 \\ \hline
         Jie Jian & 689 \\ \hline
    \end{tabular}
    
\end{table}
\subsection{Experimental Setup and Ablation Study}
\label{sec:experimental setup}

The performance of LAND is evaluated based on three variants of the KGE models: \textbf{LAND}, \textbf{LAND-$g_{lin}$}, and \textbf{LAND-$g_{gru}$}. The first variant consists of a unimodal KGE model, i.e. DistMult, which only considers the structural relations between entities in the KG, and no literal (e.g., text or publication dates) is considered. The second variant LAND-$g_{lin}$ incorporates titles of papers into the embeddings modeled by DistMult by learning the parameters of a linear transformation. The third variant LAND-$g_{gru}$ uses numeric attributes of the nodes (e.g., publication dates) along with titles and it incorporates them into entity representations by using a GRU function \cite{cho_properties_2014}. The implementation of the multimodal KGE models is made compatible with \texttt{PyKEEN (v.1.4.0)} \cite{ali_pykeen_2021}. The source code of different variants as discussed previously is available on Github\footnote{\url{https://github.com/sntcristian/and-kge}}. The KGE models are trained and evaluated using Colab Pro notebooks with $\approx$ 24GB of RAM and Nvidia Tesla T4/K80 GPUs.


Two major tasks are involved in these experiments, i.e., i) an evaluation of LAND against a candidate set of authors associated with an ORCID iD in OC-782K and ii) a generalizability analysis of LAND on the benchmark dataset provided by AMiner in \cite{zhang_name_2018}, where LAND is compared to the SOTA models surveyed in \emph{Section \ref{sec:related_work}}. Inspired by~\cite{zhang_name_2018}, the evaluation metrics \textbf{pairwise Precision}, \textbf{Recall}, and \textbf{F$_1$} are used. For studying the generalizability of LAND, these metrics are macro-averaged across all 100 test names. 

In addition, we have to state the generalizability study is unfair since the other graph embedding models make use of more features because AMiner-534K does not include abstracts and keywords. The reason why we wanted to test our model on an extracted KG and compare the results with those obtained on a bigger benchmark is to test the hypothesis that our architecture maintains competitive performances even on a more challenging benchmark where the number of authors is relevantly higher.

\subsection{Model Selection and Threshold Analysis}
\label{sec:model_selection}

The models are trained using the Binary Cross Entropy (BCE) loss function, the Adam optimizer, the Stochastic Local Closed World Assumption (SLCWA) training approach, and label smoothing as a regularization technique. Note that for training, each KG is split with a ratio of 64\% training, 16\% validation, and 20\% testing. 
Random search has been used to perform the hyper-parameter optimizations over the range of values given in \emph{Table \ref{tab:hpo}}. Each model is trained for a maximum of 1000 epochs and early stopping is applied to speed up the optimization process and avoid overfitting.

\begin{table}[]
    \centering
     \caption{Hyper-parameter ranges for the HPO studies.}
    \label{tab:hpo}
    \begin{tabular}{|c|c|}
    \hline
          Hyper-parameters & Ranges \\ \hline
Embedding dimension	& {128, 256, 512} \\ \hline
Learning rate (log scale) &	[0.0001, 0.01) \\ \hline
Number of negatives per triple (log scale) &	[1, 50) \\ \hline
Batch size	& {128, 256, 512} \\ \hline
Smoothing coefficient (log scale) &	[0.001, 1.0) \\ \hline
    \end{tabular}
\end{table}

Note that due to the limitation of resources, we ran the optimization study only for the unimodal model (i.e., DistMult) on both datasets and chose the set of optimal hyperparameter values which gave the best results, and then decided to apply them also for training the multimodal models.  The optimal hyper-parameters are as follows: for OC-782K, embedding dimension: 512, learning rate: 0.0003, number of negatives: 12, batch size: 512, smoothing coefficient: 0.001, epochs: 120; for AMiner-534K, embedding dimension: 128, learning rate: 0.0001, number of negatives: 32, batch size: 512, smoothing coefficient: 0.1, epochs: 300.

For HAC, we define the distance threshold for the final clusters experimentally by trying to find a trade-off between Precision and Recall, by finding the threshold which gives the best F$_1$ score. However, since high recall systems tend to group different authors together and this negatively affects the performances for AND, we decided to favor high precision over recall. For OC-782K, the resulting best threshold is 0.6, while for AMiner-534K it is 0.26.

\subsection{Baseline Methods}

To better assess the performances of the LAND framework, two baseline methods are implemented: (i) a rule-based method originally proposed in \cite{caron_large_2014}, which assigns a pairwise score of similarity to two publications based on several rules; and (ii) a simple disambiguation algorithm based on blocking and clustering of sentence embeddings extracted from titles. 

We selected the rule-based method inspired by \cite{caron_large_2014}, hereby mentioned as \emph{Score-Pairs}, for three main reasons: (\emph{i}) for its simplicity and scalability; (\emph{ii}) for the fact that it was recently used effectively on a SKG in \cite{farber_microsoft_2022}; (\emph{iii}) for the fact that it has an open implementation on GitHub\footnote{\url{https://github.com/lin-ao/enhancing_the_makg}}. Moreover, other methods surveyed in \emph{Section \ref{sec:related_work}} present issues in reproducibility, since most of them do not have an open-source implementation, some of them are supervised or they rely on many hidden parameters that need to be tuned for OC-782K.

Score-Pairs classifies if two publications belong to the same author or not by looking at several features (e.g., shared words in title, co-authors, citations, etc.) and computes an affinity score for each one of these features based on a list of criteria, i.e., exact string matching or the number of co-occurrences. A list of the features compared, along with the respective comparison criteria and scores are reported in \emph{Table \ref{tab:caron}}. Then, a threshold on the sum of the affinity scores is chosen in order to decide whether the publications, given the similarity of their attributes, belong to the same author or not. In our experiments, the value of the threshold is 10.

\begin{table}[]
    \centering
       \caption{Rules to compute the similarity of two pairs of publications for the baseline Score-Pairs. This table is a subset of the rules originally introduced in~\cite{caron_large_2014} }
    \label{tab:caron}
    \begin{tabular}{|c|c|c|}
    \hline 
         Field  & 	Criterion &	Score \\ \hline
Shared words in titles  & 	1 / 2 / $>$ 2  & 	3 / 5 / 8  \\ \hline
Shared coauthors  & 	1 / 2 / $>$ 2	  & 4 / 7 / 10 \\ \hline
Journal & 	Exact match	 & 6 \\ \hline
Shared cited works & 	1 / 2 / 3 / 4 / $>$ 4  & 	2 / 3 / 6 / 8 / 10 \\ \hline
Self-citation & 	one publication citing the other  & 	10 \\ \hline
    \end{tabular}
 
\end{table}

The second baseline \emph{Title-Similarity} is chosen to estimate the representativeness of textual embeddings for the task of AND. This baseline performs HAC on the title embeddings encoded by the SPECTER language model presented in \cite{cohan_specter_2020} and can be looked at as a variation of our architecture which uses only title embeddings. The results of this baseline are reported to test the influence of using structural information of a KG over just textual information; which, in the case of titles, might be not so much discriminative.

The clustering algorithm of Title-Similarity is implemented as follows: single linkage as linkage method, cosine similarity as affinity measure, and a threshold of 0.18. As for the architecture using KGEs, the threshold for clustering is selected by maximizing the F$_1$ score while favoring Precision over Recall. 

In the generalizability study, our architecture was compared with several baselines tested on the AMiner benchmark \cite{zhang_name_2018}, including the graph based model \cite{fan_graph-based_2011}, the supervised learning model \cite{ngonga_ngomo_ethnicity_2016}, and the graph embedding models \cite{zhang_name_2017}, \cite{zhang_name_2018}, \cite{wang_author_2020}, \cite{pooja_exploiting_2021} and \cite{chen_name_2021}. A description of each of these models is provided in the respective subsections of \emph{Section ~\ref{sec:related_work}}.

\subsection{Results}

\paragraph{Evaluation on OC-782K}

This section compares the results of different LAND variants, i.e., DistMult + HAC, DistMultLiteralE-$g_{lin}$ + HAC and DistMultLiteralE-$g_{gru}$ + HAC, on OC-782K with the two previously described baseline models. 

\begin{table}[]
    \centering
        \caption{Results of AND on OC-782K. The best results are highlighted in bold.}
    \label{tab:resultoc1}
    \begin{tabular}{|c|c|c|c|}
    \hline
         Model  & 	Precision  & 	Recall  & 	F$_1$ \\ \hline
Score-Pairs & 	84.66	 & 50.20	 & 63.03 \\ \hline
Title-Similarity & 	71.56 & 	66.64	 & 69.02 \\ \hline
LAND & 	{\bf 91.71}	 & \underline{67.11} & {\bf 77.50} \\ \hline
LAND-$g_{lin}$ & 	89.63 & 	66.98 & 	76.67 \\ \hline
LAND-$g_{gru}$ & 	82.76 & 	{\bf 67.59} & 	74.40 \\ \hline
    \end{tabular}

\end{table}

\emph{Table~\ref{tab:resultoc1}} shows the results of the experiments. The embedding-based models outperform the two baseline methods except for the precision of LAND$g_{gru}$. More precisely, there is an increment in the pairwise F$_1$ score of the best performing model LAND, i.e., more than 14\% and 8\% as compared to the baselines Score-Pairs and Title-Similarity respectively. The best precision of \textbf{91.71} and the best F$_1$ score of \textbf{77.50} is obtained by LAND. The best recall is {\bf 67.59} obtained by LAND-$g_{gru}$. However, the difference in the recall as compared to LAND is marginal. 

The improvements over the baseline Score-Pairs show that KGE embeddings are capable of modeling very discriminative representations if compared with the rule-based method \cite{caron_large_2014} and \cite{farber_microsoft_2022}. Indeed, the cost of using a representation learning architecture that requires training on unlabelled data is repaid by the fact of obtaining more efficient features to cluster for AND, given the same amount of information, i.e. titles, coauthors, citations, venues, and publication dates.

Moreover, the low precision of Title-Similarity if compared to our LAND variants shows how beneficial is considering the whole information from the KG, with an increment in Precision of 28,16\% if compared to our most precise model.

For the multimodal models LAND$g_{gru}$ and LAND$g_{lin}$, the absence of improvements in the results is surprising and they show that for datasets belonging to one specific domain (i.e. Scientometrics), literal information does not add much information to the embeddings but rather introduces some noise.

\begin{table}[]
    \centering
     \caption{Confusion matrix of LAND on OC-782K with high precision setup. }
    \label{tab:resultoc2}
    \begin{tabular}{|c|c|c|c|}
    \hline
           &		Positive label &	Negative label &		Total \\ \hline
Positive Classification &		996 &		90 &		1086 \\ \hline
Negative Classification &		488 &		1582 &		2030 \\ \hline
Total &		1484 &		1672 &		3110 \\ \hline
    \end{tabular}
   
\end{table}

\begin{figure}
    \centering
    \includegraphics[width=7cm]{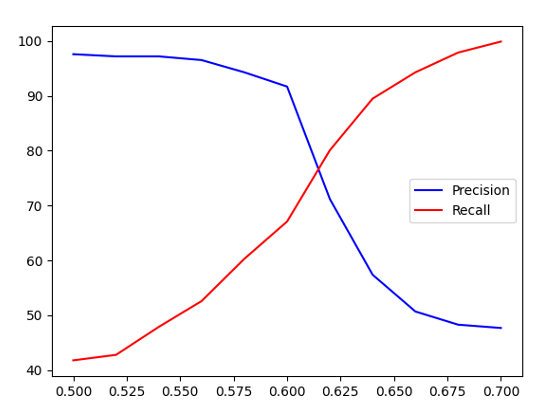}
    \caption{The plot of the precision and recall curves of our best AND system on different distance thresholds.}
    \label{fig:plot_oc}
\end{figure}

As it is noticeable in \emph{Table \ref{tab:resultoc2}}, the performances of LAND with respect to recall are far from being optimal, since our models ignored a relevant number of matching authors ($>$ 30\% avg.) in the evaluation dataset. However, we decided to avoid higher thresholds in order to reduce the number of false positives produced by our clustering algorithm and, as a consequence, to avoid attributing papers written by different persons to the same author. A plot of Precision and Recall curves for OC-782K is available in the \emph{Figure \ref{fig:plot_oc}}.


By applying LAND to the whole set of authors in OC-782K with the high precision setup, we are able to reduce the author entities from 188,565 to 135,325 (a reduction of more than 28\%). This shows how relevant KGEs can be for AND on SKGs and how they can be effective in removing duplicates.

\paragraph{Evaluation on AMiner Dataset}

We tested the generalizability of our approach on a newly collected KG extracted from the AMiner benchmark dataset for AND~\cite{zhang_name_2018}. The results of LAND are compared to the performances of SOTA AND models reported in the benchmark study \cite{zhang_name_2018} and with other graph embedding models which were tested on the AMiner benchmark (a description of these models is provided in \emph{Table \ref{tab:table_graph_and}}). However, we remark that this comparison is not entirely fair, due to the lack of certain information, e.g., abstract and keywords, in AMiner-534K. For more details on why we did not add this information see \emph{Section \ref{sec:introduction}} and \emph{Section \ref{sec:dataset}}. The results of \cite{fan_graph-based_2011}, \cite{ngonga_ngomo_ethnicity_2016}, \cite{zhang_name_2017}, \cite{zhang_name_2018} are taken from \cite{zhang_name_2018}. The other results are taken from the respective papers.

\begin{table}[]
    \centering
        \caption{Results of author name disambiguation for the AMiner benchmark \cite{zhang_name_2018}. The best results are reported in bold and the underlined results show the second-best result.}
    \label{tab:resultaminer}
    \begin{tabular}{|c|c|c|c|}
    \hline
         Model  & 	Precision  & 	Recall  & 	F$_1$ \\ \hline
\cite{fan_graph-based_2011}	 & \underline{81.62}	 & 40.43	 & 50.23 \\ \hline
\cite{ngonga_ngomo_ethnicity_2016}	 & 57.09	 & 77.22	 & 63.10 \\ \hline
\cite{zhang_name_2017} & 	70.63 & 	59.53 & 	62.81 \\ \hline
\cite{chen_name_2021} & 	65.59 & 69.96 & 65.71 \\ \hline
\cite{pooja_exploiting_2021} & 62.6 & \textbf{76.1} & 66.9\\ \hline
\cite{zhang_name_2018} & 	77.96 & 63.03 & \underline{67.79} \\ \hline
\cite{wang_author_2020} & \textbf{82.23} & \underline{67.23} & \textbf{72.92} \\ \hline
LAND & 	78.36 & 	59.68 & 	63.36 \\ \hline
LAND-$g_{lin}$ & 	77.24 & 	61.21 & 	64.18 \\ \hline
LAND-$g_{gru}$ & 	77.62 & 	59.91 & 	63.07 \\ \hline
    \end{tabular}

\end{table}

LAND-$g_{lin}$ outperforms other SOTA models which do not use embedding methods, such as \cite{fan_graph-based_2011} and \cite{ngonga_ngomo_ethnicity_2016}. This shows how encoding structural information from linked data with knowledge graph embeddings is more effective than supervised methods or predefined heuristics, even when we do not use abstracts and keywords. Moreover, it's interesting to see how LAND-$g_{lin}$ also outperformed \cite{zhang_name_2017}, which is a graph embedding method based on coauthorship information, suggesting the advantage of our architecture which models also relations between publications and venues, and relations between authors and affiliations.

However, as a negative result, we see that our method is outperformed by other graph embedding techniques such as \cite{zhang_name_2018}, \cite{chen_name_2021}, \cite{pooja_exploiting_2021} and \cite{wang_author_2020} in terms of F$_1$. Nonetheless, our LAND architecture which does not make use of literal achieves the second best Precision score among the graph embedding models, while the Recall is considerably lower than that of the other models. However, the low recall of the model is explained by the fact that our LAND architectures are trained on a KG which does not contain information from abstracts and keywords.

Moreover, we claim that the relatively lower performances of our model are compared with the architectures of \cite{zhang_name_2018}, \cite{chen_name_2021}, \cite{pooja_exploiting_2021} and \cite{wang_author_2020} are balanced by the following advantages of our architecture: 

\begin{itemize}
    \item The approach by \cite{zhang_name_2018} require human-annotated data in the training process in order to learn structure-based features from the dataset. Instead, our LAND architecture learns discriminative features for AND from an SKG without needing human supervision.

    \item \cite{zhang_name_2018}, \cite{pooja_exploiting_2021} and \cite{chen_name_2021} make use of complex feature engineering methods in order to create scholarly networks from bibliographic metadata. This feature engineering methods include document encoding by means of Doc2Vec, document similarity estimation and coauthorship similarity estimation in order to create content or coauthorship graphs from bibliographic metadata. This process is not only time-consuming but requires many parameters to be tuned for each dataset, and this hinders the reproducibility of their approach. Instead, our approach does not need feature engineering and learns node embeddings directly from the relations made explicit in the KG.

    \item \cite{wang_author_2020} is the most similar architecture if compared to ours. The astonishing performances of this model suggest the advantages of using GANs to incorporate content information into relational information from heterogeneous graphs. However, this architecture employs two different modules, a discriminative module, and a generative module, to refine node embeddings in an adversarial fashion. Moreover, the lack of available source code makes this complex architecture difficult to reuse. Instead, our model incorporates content information along with relational information in only 1 model, i.e. DistMultLiteralE \cite{kristiadi_incorporating_2019}, of which we provide an open-source implementation.

\end{itemize}

\subsection{Error Analysis} 

We randomly sampled a subset of 50 wrongly matched pairs (i.e. false positives) from the disambiguated OC-782K in order to 
analyze the most frequent errors produced by our AND system. We found out that most of the wrong matches are related to Asian authors with common surnames and first initials, like “Chen B”, “Kim S”, “Li Y”, “Wang J”, “Li J”, “Hu Z” and “Chen J”. This is probably due to the fact that LN-FI blocking tends to create huge blocks for very frequent surnames and this causes wrong authors to slide inside the final clusters, especially when they share some features (like references or publishing venue). However, we found out that it is possible to remove all these errors by using a post-blocking strategy which poses the condition that $full name_i= full name_j$ before merging two authors. Indeed, we found out that all the wrongly matched pairs in our sample which share the same full name are the same person and their entities are wrongly labeled due to the fact that they used multiple ORCIDs across different scholarly works. However, this post-blocking strategy was not included in our framework since was tested only on a small number of false positives and it is limited to the availability of at least a last name and a full name for each author reference, which is not always the case, especially for old publications.
\section{Summary \& Future Perspectives}
\label{sec:conclusion}
This article has introduced a framework, named LAND, to perform Author Name Disambiguation (AND) for scholarly data represented as linked data or included in SKGs by developing KGE models based on relationships between entities and the related literal information associated with them. We have demonstrated that these models can be used in the downstream task of clustering for AND effectively. The proposed framework outperforms state-of-the-art methods on
a newly created benchmark dataset defining an SKG (named OC-782K) compliant with the OpenCitations Data Model (OCDM) as well as another SKG (named AMiner-534K) created using an existing benchmark dataset, i.e., AMiner. Our method is able to maintain competitive levels of precision, recall and F$_1$ even when dealing with more complex models. 
Moreover, LAND is designed for dealing with data within knowledge graphs.

In the future, we plan to extend our approach to include also author collaboration network information along with the topic of interest/area of expertise extracted by processing the author's publications via deep learning approaches. Having such additional data will allow us to test if they can improve the results for the task of author name disambiguation.

\section*{Acknowledgments}

This study was partially funded by the ``Scholarship for research periods abroad aimed at the preparation of the master thesis" by the department of Classical Philology and Italian Studies, University of Bologna (\url{https://ficlit.unibo.it/it}).

\bibliographystyle{unsrtnat}
\bibliography{biblio}  






\end{document}